Regular Article

# Brain over Brawn: Using a Stereo Camera to Detect, Track, and Intercept a Faster UAV by Reconstructing the Intruder's Trajectory

**Antonella Barišić**⬡, **Frano Petric**⬡ and **Stjepan Bogdan**⬡
Laboratory for Robotics and Intelligent Control Systems, Faculty of Electrical Engineering and Computing, University of Zagreb, Unska 3, Zagreb, Croatia

**Abstract:** This paper presents our approach to intercepting a faster intruder UAV, inspired by the MBZIRC 2020 Challenge 1. By utilizing *a priori* knowledge of the shape of the intruder's trajectory, we can calculate an interception point. Target tracking is based on image processing by a YOLOv3 Tiny convolutional neural network, combined with depth calculation using a gimbal-mounted ZED Mini stereo camera. We use RGB and depth data from the camera, devising a noise-reducing histogram-filter to extract the target's 3D position. Obtained 3D measurements of target's position are used to calculate the position, orientation, and size of a figure-eight shaped trajectory, which we approximate using a Bernoulli lemniscate. Once the approximation is deemed sufficiently precise, as measured by the distance between observations and estimate, we calculate an interception point to position the interceptor UAV directly on the intruder's path. Our method, which we have significantly improved based on the experience gathered during the MBZIRC competition, has been validated in simulation and through field experiments. Our results confirm that we have developed an efficient, visual-perception module that can extract information describing the intruder UAV's motion with precision sufficient to support interception planning. In a majority of our simulated encounters, we can track and intercept a target that moves 30% faster than the interceptor. Corresponding tests in an unstructured environment yielded 9 out of 12 successful results.

## 1. Introduction

As unmanned aerial vehicles become more capable and available, there is a rise in discussion concerning the regulating policies surrounding such systems (Fox, 2019) and the overall safety and security of UAVs (Yaacoub et al., 2020), (Best et al., 2020). The most common approach to counter intruding UAVs is jamming the radio signal (Abughalwa et al., 2020), but with the rapid development of autonomous UAV solutions that do not require operator input, those techniques become increasingly obsolete, with the alternative being active defense by interceptor UAVs. Such a scenario is considered in the Mohamed Bin Zayed International Robotics Challenge 1 (MBZIRC, 2020). The Challenge 1 consists of two tasks: intercepting an intruder UAV and removing balloons from stationary tethers. In this paper, we focus on the first task, shown in Figure 1, since it is closer to real-world applications.

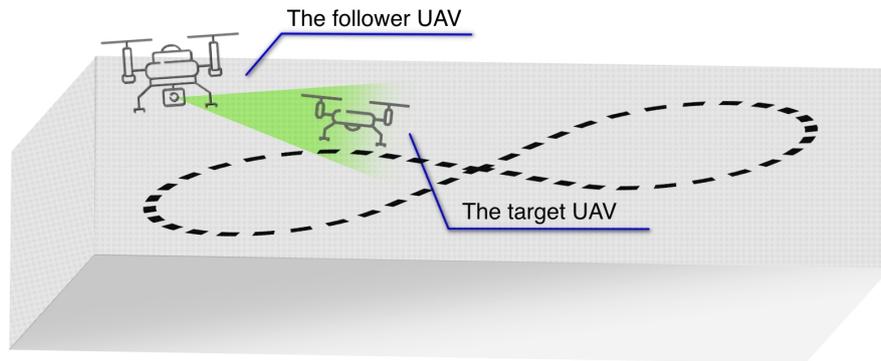

**Figure 1.** The scenario of an intruder UAV subtask of the MBZIRC2020 Challenge 1. The target UAV is following a 3D trajectory in the shape of a figure-eight with variable location and orientation inside the arena of size 100 m × 60 m. The interceptor UAV must autonomously track and detach a soft target from the target UAV, effectively mimicking the interception of the intruder UAV.

Intercepting a runaway UAV is a well-researched topic in the aerial robotics community (Beard et al., 2002), (Moreira et al., 2019), (Hehn and D'Andrea, 2012), but Challenge 1 presents more structure than a simple "runaway UAV" scenario. This simplification arises from the fact that the target's trajectory shape is known in advance, while other parameters, such as size, location, and, more importantly, orientation, are not. In the competition's previous iteration (MBZIRC2017), the fully cooperative target moved on the ground and in a predefined path. In the 2020 edition the challenge is more difficult since the target is flying in 3D space, but the target is also not fully cooperative, since all the parameters of its motion are not known in advance. Even though Challenge 1 is not tackling a general case of an intruder UAV, one can envision a scenario in which the intruder repeats a more-or-less regular pattern to linger over a specific area. In real-world scenarios, this idea can be applied to the security of airports, nuclear power plants, military buildings, prisons, and other high-security areas.

In this paper, we present a vision-based system designed to intercept intruder UAVs in an unconstrained environment. Our fully autonomous approach requires no input from the operator. Since the appearance of the intruder is not known in advance, we use a deep-learning-based object detector to overcome this problem and extract actionable information about the target. Our system processes stereo depth data through a custom, histogram filter and then, in combination with the detection results, calculates the target's 3D position. Continuous and robust information over time is provided by a tracking-by-detection algorithm based on the Kalman filter. Based on the observed target positions, we approximate the target's trajectory with a Bernoulli lemniscate and select an interception point on a straight section of the figure-eight pattern.

In the next section, we present the contributions of this work and position our research with respect to the state of the art in the field. The system architecture is described in Section 3, while in Section 4 we give a detailed description of the method used to determine the 3D position of the target object. Section 5 is dedicated to the presentation of an estimator of figure-eight shaped trajectory. Visual servoing and the target interception procedure are described in Section 6. Finally, experimental results are presented and commented in Section 7, and concluding remarks are given in the last section of the paper.

## 2. Contributions and related work

In general, UAV counter-action includes three activities: detecting, tracking, and interdicting (Guvenç et al., 2018). Detection, as the first step in the process, relies on techniques that include identification of i) RF signals from remote controllers (Ezuma et al., 2019), ii) acoustic footprint of propellers (Dumitrescu et al., 2020), iii) reflections obtained from a low-cost radar, or iv) images obtained by optical sensors (cameras). In some applications, a combination of those techniques is used to improve the probability of UAV detection. In most of the methods that employ RF and acoustic signals, as well as low-cost radars, the sensors are spatially distributed over an area of interest (e.g., an airport) so that some algorithm, based on signal time-of-arrival differences, can be applied. On the other hand, optical sensors are typically movable, mounted on a UAV that patrols over an area that should be protected. Recent detection methods (video stream processing) for such scenarios are predominantly based on the artificial intelligence paradigm - in most cases, deep neural networks are at the core of the method (Çetin et al., 2020). In general, deep learning based object detection methods are categorized into two-stage and one-stage detectors. Due to the strong emphasis on detector efficiency in UAV applications, one-stage detectors, such as YOLO (Redmon and Farhadi, 2018) and SSD (Liu et al., 2016), and lightweight networks, whether new architectures or modifications of milestone detectors (e.g., YOLOv3 Tiny), are better choices. A promising trade-off between accuracy and efficiency is also offered by recent work on anchor-free one-stage detectors, like CenterNet (Duan et al., 2019) and FSAF (Zhu et al., 2019).

Once detected, a target UAV is usually tracked by the same technique used for detection, or alternatively, combination of sensors fixed on the ground and sensors mounted on the tracking UAV(s) can be implemented. Deploying tracking UAV(s) requires estimation of the trajectory of the target UAV and calculation of the tracking path. Such scenario, if successfully applied, finally ends in interdiction of the target by a single tracking UAV (for example by an on-board tethered net system (AeroGuard, 2020)) or by a group of UAVs (Brust et al., 2021).

The goal of the MBZIRC2020 Challenge 1 was to detach a ball suspended from the target UAV, prompting many teams to opt for detecting the ball rather than the UAV. In this paper we aim for a more general use-case and focus on the UAV, more specifically on the first part of the challenge: detecting and intercepting an UAV, by building upon our previous work (Barisic et al., 2019). To detect and track the target UAV, we use the stereo camera ZED Mini and the deep neural network YOLOv3 Tiny (Redmon and Farhadi, 2018) trained on our own dataset of 13,000 images. This differs from most of the published approaches from the MBZIRC2017 and the large body of research where the detection of the target is based on markers (Tzoumanikas et al., 2018), (Beul et al., 2019), (Li et al., 2018). To contribute to the research community, we publicly release our validation dataset named UAV-Eagle. The UAV-Eagle dataset provides a benchmark in object detection of UAVs in an unconstrained environment characterised by illumination changes, motion effects, viewpoint changes, and a high-density environment.

Combining UAV detections with depth estimation from ZED, we reconstruct 3D position of the target and feed it to a Kalman filter to achieve robust target tracking. A similar approach using an Intel RealSense D435 was reported in (Vrba et al., 2019), where the authors use a custom depth image processing based on classical computer vision methods to detect intruder UAVs with the assumption that there are no other flying objects in the field of view. A deep neural network approach to a similar task was published in (Vrba and Saska, 2020) in which the authors assume a known size of the target to reconstruct its position. Alongside being more general compared to the most recent research, this work is backed by a more powerful graphics processor and a stereo camera with a larger baseline, providing improved performance. On top of the CNN-based UAV detection, we develop a depth processing algorithm that takes into account that the UAV is an object with known structure to generate more consistent and robust depth estimation in outdoor conditions.

To the best of our knowledge, all current interception approaches require the interceptor to be faster than the intruder (Moreira et al., 2019), some even up to two times faster than the target (Yang and Quan, 2020), which will be increasingly difficult to achieve, especially since some researchers in the field estimate that within 5 years there could be UAVs with a top speed of 100 m/s (Bond et al., 2019). In this work, inspired by the challenging target speeds of MBZIRC2020 and constrained by limitations of our hardware, we explore the possibility of intercepting a faster and more agile target with a slower UAV. To that end, as a main contribution of this paper, we leverage the knowledge of the shape of the target trajectory to reconstruct the target trajectory in the global coordinate frame from sparse observations of the target in the image. Following a successful reconstruction of the target trajectory, measured by the distance between sets of observed positions of the target and the idealistic approximation of the trajectory, we select the interception point which will allow us to intercept the target with significantly reduced effort in control inputs.

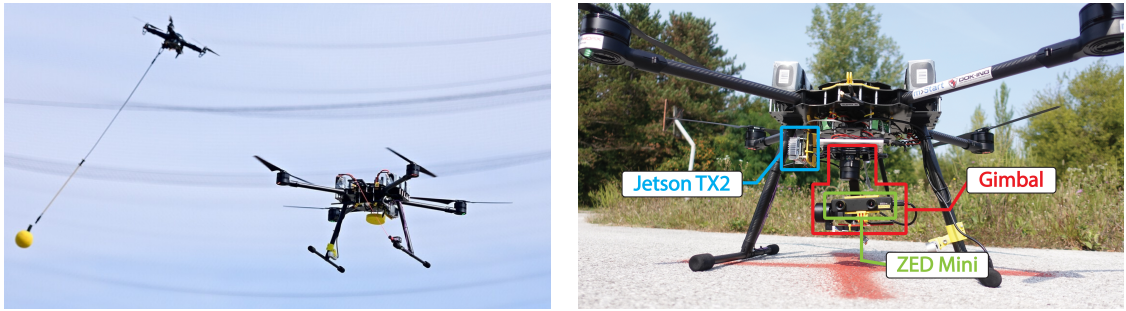

**Figure 2.** Target intercept scenario at MBZIRC 2020 challenge (left, courtesy of http://mbzirc.com/) and custom made Kopterworx Eagle UAV used by LARICS team at the competition (right).

## 3. Kopterworx Eagle UAV

The frame of our interceptor UAV (Figure 2) consists of four arms, body and two legs with skis, all built from carbon fiber. The vehicle is equipped with T-Motor Flame 60A 12S ESC speed controllers that drive U8 Lite Kv150 12S motors with 0.56 m carbon fiber propellers. The vehicle's autopilot is a Pixhawk 2.0 running ArduPilot software. The maximum take-off weight of the vehicle is 12 kg with 2 kg of payload. The vehicle is equipped with an Intel NUC i7/16GB computer running Ubuntu 18.04 LTS and ROS Melodic. The modular design of the Eagle frame allows for various computational and sensory configurations, including multiple cameras and even 3D LiDAR sensors. In the configuration used for the first challenge of MBZIRC 2020, image processing and stereo reconstruction is performed using Nvidia Jetson TX2, with images being captured by a ZED Mini stereo camera mounted on a Gremsy Pixy F 3-axis gimbal. All components of the aerial vehicle are powered by two LiPo 12S 14000 mAh batteries, giving the vehicle up to 30 minutes of flight time.

### 3.1. Software architecture

The software components can be divided into four main modules: visual perception, estimation of figure-eight shaped trajectory, global state machine and control algorithms. An overview of the software architecture is shown in Figure 3 where the main modules are highlighted with blue, red, green and orange color, respectively. All algorithms communicate using Robot Operating system (ROS). Besides object detection and depth estimation, which are performed on the Jetson TX2, all other software components are running on the on-board Intel NUC computer.

The inputs of visual perception are color and depth images acquired from ZED Mini stereo camera. A convolutional neural network inference is performed on a given color image, while depth is estimated by analysing the histogram of depth. The pixel coordinates of detected target and the depth estimation are forwarded to the position reconstruction module in order to obtain $(x, y, z)$ coordinates in the global coordinate system of the follower. The information about the target's position is further enhanced by a Kalman filter estimation.

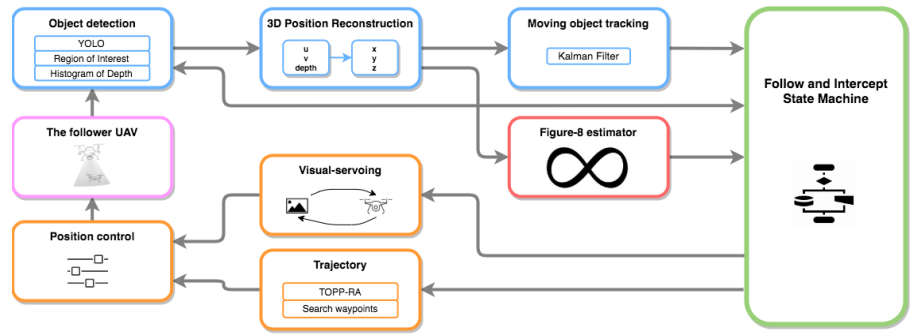

**Figure 3.** Software architecture.

The goal of moving object tracking is to keep the target in the field of view, which is achieved by a position-based visual servoing (PBVS), which is a well-researched topic in the field of robotics (Sinopoli et al., 2001), (Chaumette and Hutchinson, 2006). The visual servoing module transforms errors in relative position with respect to the target into references for the on-board UAV controller. In order to continuously follow and promptly respond to target's movement, PBVS relies on the aforementioned Kalman filter estimation. The information for estimation of the figure-eight shaped trajectory and interception point is collected during the stint in which we are able to follow the target. Once the target is lost, the interceptor switches to local search mode trying to find the target again. Local search trajectories are generated using Time-Optimal Path Parameterization (TOPP-RA) library (Pham and Pham, 2018) and follow a Levy flight paradigm (Puljiz et al., 2012). The behavior switching is managed by a state machine which implements the proposed Search-Follow-Intercept strategy, with the emphasis of this work being on the Follow and Intercept parts of the strategy, as described in the remainder of the paper.

## 4. 3D position of the target

The starting point of the proposed system is an object-detecting module. More precisely, a module that can detect multicopters of any kind. Due to the complex structure of multicopters and to achieve generalisation among different appearances of various multicopters, we apply a convolutional neural network (CNN) to solve this problem. Among the advanced object detectors, YOLOv3 was selected as the best solution due to its high efficiency and good accuracy, as confirmed by many researchers. Specifically, we used a lightweight version of the network, YOLOv3 Tiny, which we modified by adding another YOLO layer to perform detection across three scales of feature maps to improve the detection of objects that occupy only a small portion of the image. The network architecture consists of 30 layers, of which 16 are convolutional layers. In the absence of a publicly available dataset of multicopters, we collected numerous images from the internet and filtered out duplicates and outliers using an open-source clustering method based on image fingerprints generated by a pre-trained, deep convolutional network. After manual and pseudo-labeling, the final result is an annotated dataset of 13,000 images of various multicopters in different environments. In comparison to our previous work (Barisic et al., 2019), we extend our dataset with unlabeled images of objects that have a similar appearance to UAVs to achieve more robust detection.

Before training, anchor boxes used as priors for the prediction of bounding boxes are calculated by the K-means clustering algorithm on our dataset. The training started from the pre-trained weights from the COCO dataset and the following training parameters were used: batch size = 64, momentum = 0.9, decay = 0.0005 and learning rate = 0.001. The network was trained until the classification accuracy stopped improving, and the weights at 650 000 iteration were selected as the best based on the mean average precision (mAP) on the training and validation dataset to avoid overfitting. As presented in Figure 4, the trained network is able to detect different types of multicopters such as DJI Phantom 4, our custom aerial platform Eagle and Gazebo models of ArduCopter 3DR and Eagle. The presented results show the robustness of the trained YOLOv3 network for different weather conditions (foggy and sunny), different backgrounds and for both small and large objects.

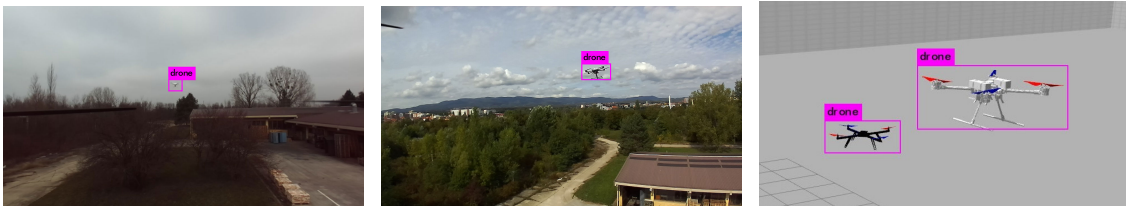

**Figure 4.** Object detection of multicopters used as targets for experiments. The target on the left image acquired from on-board ZED Mini stereo camera is a DJI Phantom 4. Even though barely seen by the human eye, the trained CNN successfully detects distant objects in difficult weather conditions. The center image shows the detection at the test site of our custom aerial platform Kopterworx Eagle. Despite the fact that the network has never seen the image of this multicopter during training, it is able to detect this target with high accuracy. The detection results of ArduCopter 3DR and aerial platform Eagle models in Gazebo simulator are shown on the right image.

An output of CNN inference is a list of detected objects defined by their position coordinates and dimension in the image plane. As we expect only one target object, the data association is performed based on the Kalman filter with the constant velocity model for a single instance. In the case of multiple detections, the measurement with the highest value of Intersection over Union (IoU), compared to the current estimate of the filter, is considered the most appropriate. Even though it is unlikely that multiple detections will occur in a single frame, the resilience to false positives is improved by enforcing the data association.

Following several consecutive detections of the target, we apply the Region of Interest (RoI) concept. The RoI is defined as the bounding box of the previous detection expanded equally in all directions by a certain factor. If there is no detection in the current frame, an estimation of the bounding box from the Kalman filter is used to define RoI. The minimum size of RoI is specified as the input size of the CNN, which is $608 \times 608$ pixels. The objective of the RoI is to keep only the part of the image where the target is expected to be found. In such manner, information from previous frames is exploited and detection of small objects is improved. If there are no new detections in the previous frames, the RoI is deactivated and another set of consecutive measurements is required to activate it again.

### 4.1. Depth of a well structured object

Calculating the image-depth of a well structured object based on the detection in the image, i.e., the bounding box, is not trivial, because there is no prior knowledge of which pixels in the bounding box are occupied by the real object. Under perfect conditions, such as a simulation setup, the depth could be obtained by averaging the depth values of all the pixels in the bounding box. However, field experiments with a real sensor have shown that, in the case of multicopters, depth measurements are noisy. This problem can be partially eliminated by using a gimbal-mount to minimize camera motion. The most challenging situation is the one with a multicopter without an outer shell, such as Eagle, because measurement values can vary due to the 'gaps' in the UAV body. Additionally, a large distance from the target, as well as simultaneous motion of the sensor and the target, aggravate the measurement errors.

Given all the above, taking into account all measurements enclosed by the bounding box results in noisy and unreliable depth data. As a solution, we propose to distribute the depth data from the bounding box across bins of histogram in order to separate the true data from noise and inaccurate measurements. The peaks in histogram, i.e., the bins with the number of measurements higher than its two neighbouring bins, one on each side, are selected as candidates. Two presumptions are made for selecting a final value: i) there are no other objects between the sensor and the target, and ii) the measurements contain more true depth values than noise. According to the second presumption, candidates are narrowed down to peaks that have a higher or equal number of measurements than the average number of measurements of all peaks. By the first assumption, no other objects lie between sensor and target, so the average value of the first candidate peak constitutes the best depth estimate. A good example of advantages of the proposed approach is shown in the right image in Figure 5. Obtained measurements extend in the range of 6 m to 15 m, which can by no means be accurate because diagonal from motor to motor of the target is 1.13 m. The corresponding histogram of depth contains eight peaks, which are reduced to only one after applying the second assumption.

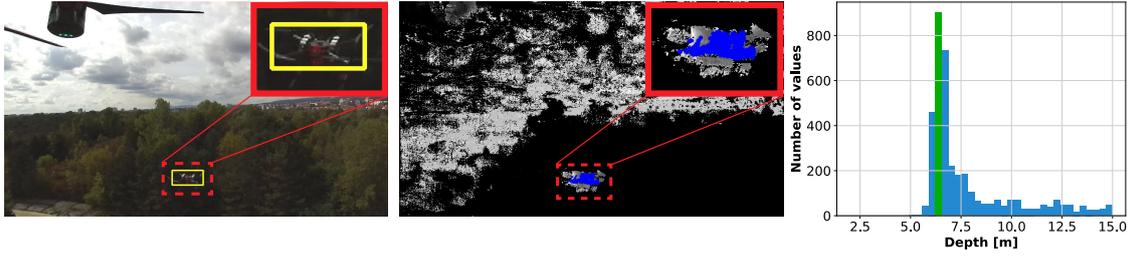

**Figure 5.** An example of extracting relevant depth information from noisy and inaccurate measurements by analysing the depth histogram (right image). Black pixels in the depth image (middle image) represent pixels where the depth could not be determined (e.g., occlusions) or where the objects are out of measurable range. The enlarged area shows the depth of pixels enclosed by the bounding box from object detection (left image), and the blue pixels are the ones from the selected bin (highlighted in green).

The maximum measurable range of depth for ZED Mini camera is from 0.1 m to 20 m. However, since the value of the measurable range affects demand for computational power and GPU memory, and also since the measurements significantly deteriorate for large distances from the target, a range from 2 m to 15 m is chosen as adequate for our application. For this range, the number of bins was experimentally determined to be 40. When the depth exceeds the measurable range, we compare the bounding box size to the image size. Based on this, we set either the maximum or minimum depth value to be output of our algorithm.

### 4.2. Calculating and tracking 3D position of the target

The final step of the visual-perception module is transformation of visual information into position information in the coordinate frame of the follower. Coordinate frames of the scenario are shown in Figure 6. By knowing the depth $d$ and target's coordinates $u$ and $v$ in the image plane, the 3D position $\mathbf{p}_m^C = (x_c, y_c, z_c)$ in camera coordinate frame $C$ can be calculated as

$$x_c = \frac{u - c_x}{f_x}d, \quad y_c = \frac{v - c_x}{f_y}d, \quad z_c = d, \tag{1}$$

where $f_x$ and $f_y$ are the focal length in pixels, $c_x$ and $c_y$ are the principal point. As the camera is fixed to a gimbal, a known transformation $\mathbf{T}_F^C$ from camera frame $C$ to the local follower frame $F$ is applied to get a relative target position $\mathbf{p}_m^F = (x_f, y_f, z_f) = \mathbf{T}_F^C \mathbf{p}_m^C$ with respect to the follower. On-board localisation sensors compute pose of the follower in global frame $G$, which can be written in a from of transformation matrix $\mathbf{T}_G^F$. Knowing the position and orientation of the follower in $G$, the target position $\mathbf{p}_m^G$ in global coordinate system $G$ is calculated by:

$$\mathbf{p}_m^G = \mathbf{T}_G^F \mathbf{T}_F^C \mathbf{p}_m^C \tag{2}$$

As position controller operates at a frequency of 50 Hz, the discrete Kalman filter with a constant velocity motion model is applied to meet the rate of the controller. The target is tracked in coordinate frame $G$, with state vector defined as $\hat{\mathbf{s}}_k^G = \begin{bmatrix} x_k^G & y_k^G & z_k^G & \dot{x}_k^G & \dot{y}_k^G & \dot{z}_k^G \end{bmatrix}$. The correction step uses observations of target position in the global coordinate system. Tracking the target position in global frame eliminates the impact of follower movement on the relative references for visual servoing, as they are obtained by converting the filter estimate back to the local frame $F$.

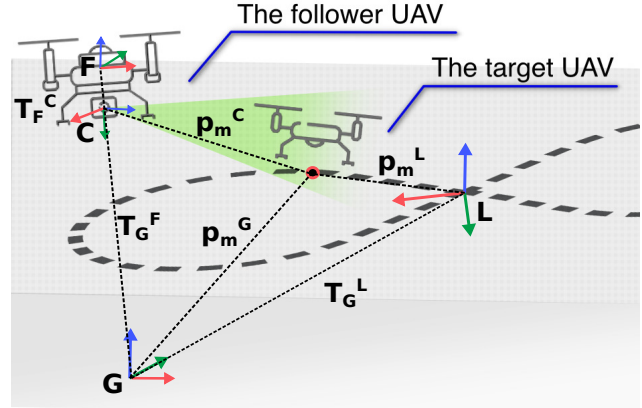

**Figure 6.** Coordinate frames and transformations used to transform measurements of target position in camera frame $C$ to a follower coordinate frame $F$ and global coordinate frame $G$ for tracking of the target and estimation of the parameters of the trajectory.

## 5. Estimator of figure-eight shaped trajectory

As stated in the description of the MBZIRC 2020 Challenge 1, the target is following a 3D trajectory in a shape of a figure-eight with a variable orientation in space. In mathematical terms, a figure-eight shaped curve is called a lemniscate. Among different representations of the lemniscate, the Bernoulli lemniscate was selected as the most suitable because of its smoothness in curvature, but also because it is characterized by a single parameter and has simple parametric equations.

### 5.1. Estimation of the Bernoulli lemniscate

The Bernoulli lemniscate is a plane curve defined by two focal points whose distance is parameter $a$, which is called the focal distance. Points on the lemniscate, in the Cartesian coordinate system, can be calculated by parametric equations:

$$x = \frac{a\sqrt{2}\cos(t)}{\sin(t)^2 + 1}, \quad y = \frac{a\sqrt{2}\cos(t)\sin(t)}{\sin(t)^2 + 1}, \quad t \in [0, 2\pi]. \tag{3}$$

Given an array of target observations $[\mathbf{p}_m]^G$ that are assumed to be lying on a lemniscate, our goal is to find the parameter $a^L$ and a homogeneous transformation $\mathbf{T}_G^L$ that, when applied to all of the collected points, results in the best overlap of the transformed points with a lemniscate with focal distance of $a^L$. Note that $G$ denotes the global coordinate system in which the observations are made, while $L$ denotes the local coordinate system attached to the lemniscate (see Figure 6). The translational part of $\mathbf{T}_G^L$ is calculated as a centroid of all target observations $[\mathbf{p}_m]^G$. The orientation of the lemniscate is calculated by performing Principal Component Analysis (PCA, (Pearson, 1901)) on $[\mathbf{p}_m]^G$, resulting in three principal axes. Taking into account that in the local coordinate frame of the lemniscate $L$, $x$ is the longest axis of the lemniscate, the unit vector of the first principal component is set as the $x$-axis of $L$. Given the fact that the lemniscate is a plane curve, the smallest principal component is set as the $z$-axis of $L$, choosing the orientation that points upwards. Finally, the $y$-axis of $L$ is set to form a right-handed coordinate frame. This procedure is detailed in Algorithm 1.

**Algorithm 1.** Estimation of the Bernoulli lemniscate

| | |
|---|---|
| **Input:** $[\mathbf{p}_m]^G$ | // Array of position measurements in $G$ |
| **Output:** $[\mathbf{p}_e]^G$ | // Array of sampled points on the estimated lemniscate in $G$ |
| **Output:** $\mathbf{T}_G^L$ | // Position and orientation of the lemniscate |
| **Parameters:** k | // Number of points for estimation |

1 **if** *new measurement* **then**
2     **Calculate transformation matrix**
3       $\mathbf{p} \leftarrow$ centroid of points in array $[\mathbf{p}_m]^G$
4       $\mathbf{x}_{axis}, \mathbf{y}_{axis}, \mathbf{z}_{axis} \leftarrow$ Principal Component Analysis of $[\mathbf{p}_m]^G$
5       $\mathbf{T}_G^L \leftarrow [\ \mathbf{x}_{axis}^\mathsf{T}\ ,\ \mathbf{y}_{axis}^\mathsf{T}\ ,\ \mathbf{z}_{axis}^\mathsf{T}\ ,\ \mathbf{p}^\mathsf{T}]$
6     **Calculate parameter** $a$ **of the lemniscate**
7       $[\mathbf{p}_m]^L \leftarrow \mathbf{T}_G^{L^{-1}}[\mathbf{p}_m]^G$      // Transform measurements to $L$
8       $[r]^L \leftarrow [\frac{\mathbf{p}_{m_x}^L}{|\mathbf{p}_{m_x}^L|}\|\mathbf{p}_m^L\|]$    // Calculate signed distance of each point in $[\mathbf{p}_m]^L$ (sign defined by $x$ component) from the origin of $L$
9       $d^L \leftarrow max([r]^L) - min([r]^L)$      // Calculate the length of the lemniscate
10      $a^L \leftarrow \frac{d^L}{2\sqrt{2}}$      // Calculate parameter $a$
11    **Calculate shift along** $x$**-axis in lemniscate coordinate frame**
12      $shift\_x \leftarrow \frac{max([r]^L) + min([r]^L)}{2}$      // If there is no offset, $max([r]) = -min([r])$
13    **Generate points of estimated lemniscate**
14      $[t] \leftarrow k$ equally spaced points between 0 and $2\pi$
15      **for** *i in range(k)* :
16        $[x_e]^L$ append $\left(\frac{a^L\sqrt{2}\cos t(i)}{\sin t(i)^2 + 1} + shift\_x\right)$
17        $[y_e]^L$ append $\left(\frac{a^L\sqrt{2}\cos t(i)\sin t(i)}{\sin t(i)^2 + 1}\right)$
18      $[\mathbf{p}_e]^L \leftarrow [[x_e]^L, [y_e]^L, [0]]$
19      $[\mathbf{p}_e]^G \leftarrow \mathbf{T}_G^L[\mathbf{p}_e]^L$

Knowing the rotation matrix and the translation vector of the lemniscate $\mathbf{T}_G^L$, the vector of position measurements is transformed into the lemniscate coordinate system. The focal distance is determined by finding two opposite endpoints on the lemniscate arcs in what is now 2D space. Using the distance between these two endpoints, the parameter $a^L$ can be determined, as shown in step 10 of Algorithm 1. The set of $k$ points, that defines the Bernoulli lemniscate, is calculated by parametric equations with the estimated value of the focal distance.

Although the algorithm is simple and very precise for a large number of points, when detections of the target are sparse, the accumulation of points on one side tends to introduce errors in the estimation of the lemniscate center. To address these errors, we exploit the symmetry of the lemniscate with respect to the origin and axes of the coordinate frame, and shift the center until said symmetry is achieved. Since this effect is more pronounced for the $x$-axis of the lemniscate (caused by the sign of the $y$ coordinate of a point on the lemniscate alternating with period of $\pi/2$ compared to period $\pi$ for the sign of the $x$ coordinate) we focus on the $x$-axis only and reuse the result $[r]^L$ (step 8 in Algorithm 1) from the calculation of parameter $a$. In determining the endpoints of the lemniscate, the resistance to possible outliers in the measurements is achieved by calculating $max([r]^L)$ and $min([r]^L)$ by taking the median of the several highest and lowest values of $r$ in array $[r]^L$.

## 5.2. Calculation of the interception point

For each new measurement of the target position, Algorithm 1 estimates the Bernoulli lemniscate that best describes the motion of the target using all collected points. To validate the estimation

**Algorithm 2.** Estimation of the interception point

---

**Input:** $[\mathbf{p}_m]^G$, $[\mathbf{p}_e]^G$, $\mathbf{T}_G^L$, $a^L$
**Output:** $position^G$, $orientation^G$      // Position and orientation of the interception point
**Parameters:** $threshold$

1 **def** $d_h(X, Y)$:
2
$$d_h(X, Y) = \max\{\sup_{x \in X} \inf_{y \in Y} d(x, y), \sup_{y \in Y} \inf_{x \in X} d(x, y)\}$$

3 **while** $1$ **do**
4   $d_H = \max\{d_h([\mathbf{p}_e]^G, [\mathbf{p}_m]^G), d_h([\mathbf{p}_m]^G, [\mathbf{p}_e]^G)\}$     // bidirectional Hausdorff distance
5   **if** $\frac{d_H}{a^L} < threshold$ **then**
6     identify $direction$ of sequentially obtained measurements
7     **if** $direction ==$ CW **then**
8       $t_i \leftarrow \frac{3}{4}\pi$
9       $t_t \leftarrow \frac{1}{4}\pi$
10     **else**
11       $t_i \leftarrow \frac{1}{4}\pi$
12       $t_t \leftarrow \frac{3}{4}\pi$
13     $position_i^L \leftarrow$ parametric equations for $t_i$
14     $position_t^L \leftarrow$ parametric equations for $t_t$
15     $\psi \leftarrow \arctan(y_t - y_i, x_t - x_i)$
16     $orientation_i^L \leftarrow [0, 0, \psi]$
17     $position_i^G$, $orientation_i^G \leftarrow$ transform $(position_i^L, orientation_i^L)$ with $\mathbf{T}_G^L$

---

and consequently conclude the estimation procedure, the bidirectional Hausdorff distance (Császár, 1978) is introduced as an evaluation metric. The Hausdorff distance between two discrete sets of points is the greatest of all distances from a set to the closest point in the other set. In the case of estimating a 3D trajectory, the first set are the 3D positions reconstructed from measurements of the target $[\mathbf{p}_m]^G$ while the second set are sampled points of the estimated lemniscate $[\mathbf{p}_e]^G$, generated by Algorithm 1 using parametric equations (3) with the current estimation of $a^L$ and $k = 100$ values of $t \in [0, 2\pi]$. When those two sets are compared, the Hausdorff distance represents the most mismatched point of the lemniscate. As the Hausdorff distance is not always symmetrical, the bidirectional form is used. The equations of the one-sided and bidirectional Hausdorff distance are given in steps 2 and 4 of Algorithm 2.

When the ratio of the bidirectional Hausdorff distance and the focal distance of the estimated lemniscate falls below a certain threshold[1], we conclude that new measurements will not significantly improve the estimation and proceed with identification of interception point. First, we identify the direction of the target in the lemniscate by using the timestamps of target detections. Based on this, we calculate the orientation of the interceptor so that it look straight towards the incoming target. Then we choose the interception point on the end of the near-straight part of the lemniscate, since this allows the longest time interval for possible corrections of the interception point.

## 6. Control of the UAV in Follow mode and switching its behaviors

The general aim of visual servoing is to keep the target within the field of view and at a desired offsets relative to the follower. The two main approaches to the visual servoing are image-based and position-based. In preparation for the competition image-based visual servoing was employed, but after the competition we decided to continue with the position-based approach because the more stable behaviour of the follower was achieved:

$$\begin{bmatrix} x_f(k+1) \\ y_f(k+1) \\ z_f(k+1) \end{bmatrix} = \begin{bmatrix} x_f(k) \\ y_f(k) \\ z_f(k) \end{bmatrix} + \begin{bmatrix} \cos\psi & -\sin\psi & 0 \\ \sin\psi & \cos\psi & 0 \\ 0 & 0 & 1 \end{bmatrix} \begin{bmatrix} x_t^F + x_{offset} \\ y_t^F + y_{offset} \\ z_t^F + z_{offset} \end{bmatrix} \tag{4}$$

---

[1] The threshold is obtained empirically and is adaptable to different applications. The exact numerical value mostly depends on the precision of the obtained measurements and the dimensions of the search area.

Based upon the relative target position, a follower position reference, which is forwarded to the position controller, is generated by Equation 4 where $k$ denotes the time step of the visual servoing module, $\psi$ is a yaw angle, $(x_t^F, y_t^F, z_t^F)$ is the target position in local frame $F$ and $(x_f, y_f, z_f)$ is the follower position in the global frame. In the case of using visual servoing to follow the target, $y_t^F$ in Equation 4 is set to zero because it is incorporated in the calculation of the desired yaw angle:

$$\psi_f(k+1) = \psi_f(k) + \arctan(\frac{y_t^F}{x_t^F}). \tag{5}$$

The follower can be operated with these four references, and which one will be set as active depends on the application and the camera pose. For example, in case of visual servoing in the interception point, $x_t^F$ would be set to zero while the $y$ component would be active. The position offsets are also adaptable to the application and for the target following only $x_{offset}$ is set to the desired value (in our case 7 m).

## 6.1. State Machine

The top-level state machine converts data, which are received from all modules in the system, into decisions regarding which task must be done next in order to achieve the end goal. The tasks defined in the Follow-and-Intercept state machine connect the operation of the different modules, while independent decisions are made in individual modules. A flowchart of the state machine is presented in Figure 7, where the diamond shape indicates a decision and the rectangles represent the operating modes. Each transition is conditioned by signals from the visual perception module. The operating modes are stated and briefly described below.

(i) IDLE - Initial mode. The follower is on the ground and motors are disarmed.
(ii) TAKEOFF - Mission starts with a request for an autonomous takeoff. Ardupilot mode is set to GUIDED_NOGPS, motors are armed and a takeoff trajectory reference is sent to the on-board controller.
(iii) SEARCH - After a successful takeoff, waypoints for search across the arena are generated. Search trajectory execution is immediately terminated if target detection occurs.
(iv) FOLLOW - In this state, visual servoing aims to eliminate position errors and to maintain the target at a relative distance of 7 m. The target's trajectory estimation is activated. If the target goes out of the field of view, the interceptor UAV returns to the 'SEARCH' state.

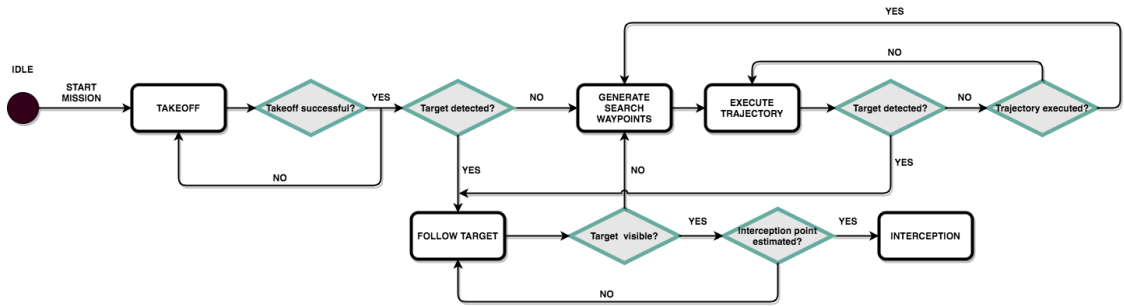

**Figure 7.** Flowchart of Follow-and-Intercept state machine.

**Table 1.** Evaluation results of the trained YOLOv3 Tiny

| | Training dataset | | Validation dataset | | UAV-Eagle dataset | |
|---|---|---|---|---|---|---|
| Number of images | 11 700 | | 1300 | | 510 | |
| mAP@0.5 | 0.9453 | 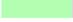 | 0.8949 | 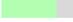 | 0.8444 | 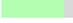 |
| Recall | 0.7448 | 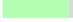 | 0.7604 | 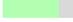 | 0.8217 | 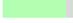 |
| F1-score | 0.8390 | 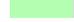 | 0.8451 | 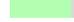 | 0.8756 | 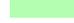 |

(v) INTERCEPT - When the quality threshold of the lemniscate estimation is reached, this state is activated, and therefore the estimated point of interception is sent to the follower as a waypoint to intercept the target.

## 7. Experimental results

### 7.1. Detection evaluation

Since our entire system relies upon its detection results, we first present a detailed analysis of the trained YOLOv3 Tiny network's accuracy and a qualitative evaluation of detection on data obtained during the MBZIRC 2020 competition. The custom dataset of 13,000 images acquired pre-competition was divided, with 90% used for training and 10% reserved for validation. As shown in Table 1, the trained network achieved mAP@0.5 value of 94.5% during training and mAP@0.5 value of 89.5% on the validation dataset. In addition to the standard validation on the training data sample, the accuracy of the trained network was also tested on the UAV-Eagle dataset, which contains 510 annotated images of the Eagle quadcopter in a challenging environment. Consequences of recording in an unstructured setting include illumination changes, motion effects, changes in viewpoint, and the presence of various background objects, such as trees, building roofs, clouds, cars, people, and so on. Therefore, the UAV-Eagle dataset is a good test of detector robustness in real-world applications and its ability to detect a previously unseen object of interest. Since the Eagle was not observed at all during the training of our CNN, the results with this dataset show a remarkable generalization of our network: mAP@0.5 of 85% on a previously unseen UAV type. The UAV-Eagle dataset is available at https://github.com/larics/UAV-Eagle.

Detecting the target UAV in the MBZIRC arena for the Challenge 1 was quite challenging for two reasons. The first reason is the large search area and small target, which means that the target often occupies only a small portion of the image, making it difficult to detect. The second reason is the resemblance of the environment to the target. The scaffolding around the arena is very similar in appearance to the airframe of the UAV, so the target UAV cannot be detected if they overlap. As can be seen in Figure 8, the trained YOLOv3 Tiny is able to successfully detect the target, even though the target occupies on average only 0.07% of the pixels of the entire image in the two sequences of detection frames shown. Two false negative detections in the case of a partially occluded target and a target that is too far away are also shown in the figure.

### 7.2. 3D position estimation

After simulation experiments that showed a root mean squared error (RMSE) of $\sim 0.05$ m and a mean absolute error (MAE) of $\sim 0.04$ m for the estimation of the target position in the global coordinate system compared to the ground truth of the Gazebo simulator, we decided to execute indoor flights with a very precise ground truth to confirm the obtained results in the real-world conditions. For this purpose, the Optitrack motion capture system with an accuracy of down to 0.2 mm was used in the test area of $6 \times 5$ m. The experiments were conducted using an AscTec Neo hexacopter as target, completing a set of three different UAVs used as targets in real-world experiments. The detection runs at frequency of 7 Hz on the Jetson TX2, as computational resources are also needed by the ZED SDK for depth estimation. The depth mode in the SDK is set to ultra and the image size is 1280 x 720.

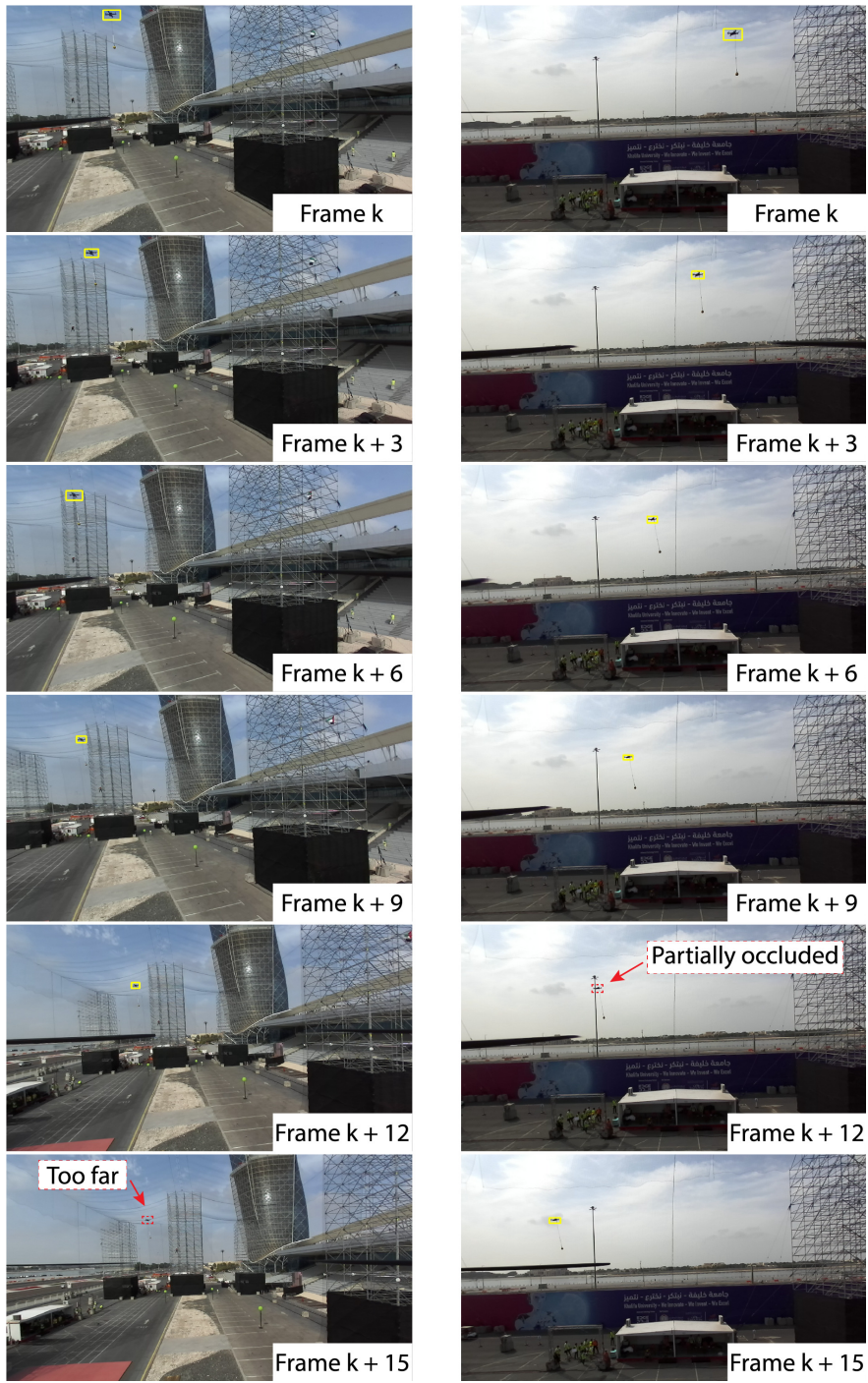

**Figure 8.** Qualitative evaluation of the trained YOLOv3 Tiny detector on the data obtained during the MBZIRC2020 competition. True positives are highlighted in yellow and false negatives in red.

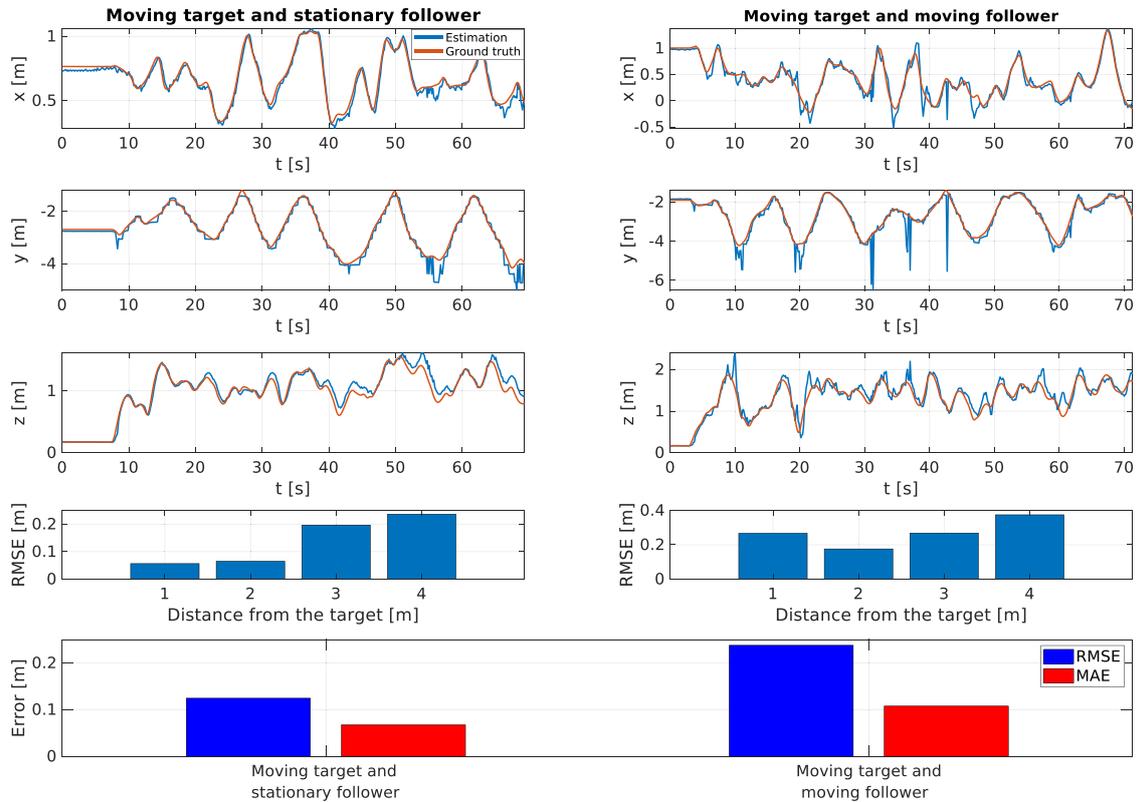

**Figure 9.** Comparison of the proposed 3D position estimation (blue) and ground truth (red) from the Optitrack system. Based on the camera motion, two different experiments are presented to show the influence of the camera motion on the value of the error. The most difficult situation for our vision-based estimation is the abrupt motion of the target. The results from both experiments are presented in terms of position over time, RMSE over distance from the target, and total RMSE and MAE.

In estimating the target position, task difficulty increases as we introduce more motion into the problem. The Challenge 1 scenario implies a moving target, and the proposed Follow-and-Intercept strategy implies a movement of the follower UAV while estimating the target position based on visual information. To show the impact of each component, the results of two experiments with different dynamic conditions are presented in Figure 9. In the first experiment, the target is moving and the camera on the follower UAV is stationary, while in the second experiment we add camera motion. In both experiments, the gap between simulation and real-world experiments can be observed as the RMSE value is significantly larger than the simulation results. This is to be expected in the real-world scenes due to the difficult imaging conditions and sensor noise. A more than twice smaller value of MAE indicates that the estimates are generally stable and occasional outliers are the result of movement. In the second experiment, the value of the errors increases because we introduced camera motion. In the presented experimental results, the sudden changes in the measurements (spikes and dips) are the result of the abrupt motion, either of the camera or of the object. These irregularities are later filtered by the Kalman procedure during tracking. It is also observed that the value of the RMSE increases as the target is further away. We consider the obtained results to be extremely good (for comparison, the diagonal from motor to motor of the target UAV in the experiments is 0.5 m) as the multicopter is a well structured object, meaning that the measured depth can be obtained from different physical parts of the target, such as motors, body frame, on-board sensors or landing gear.

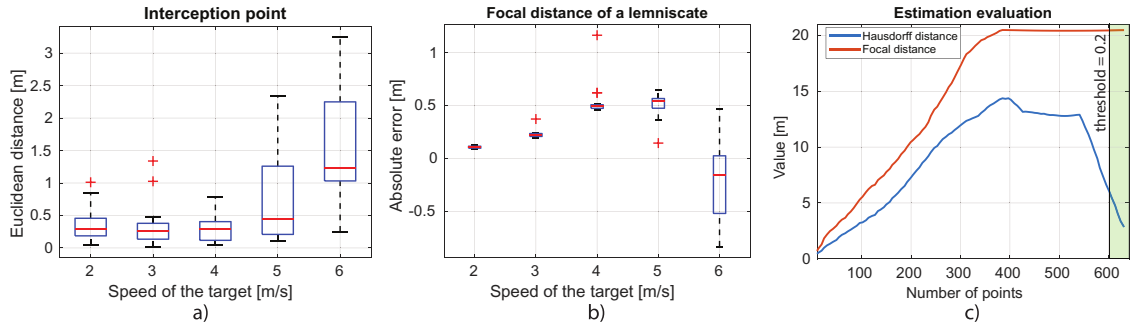

**Figure 10.** Results of the software in the loop simulation experiments for the target trajectory with a focal distance of 20 m: (a) Error of the estimated interception point in the form of the Euclidean distance from the true target position at different values of the target speed; (b) Absolute error of the estimated focal distance of the Bernoulli lemniscate at different values of the target speed; (c) The bidirectional Hausdorff distance and the focal distance over the number of acquired points of the target trajectory, at a target speed of 5 m/s.

### 7.3. Performance analysis with respect to the target speed

In order to examine the limits of the complete developed solution, we conducted a numerous simulation experiments. An important emphasis is placed on the maximum speed of the target at which we can successfully track and intercept. In the conducted experiments, the model of 3DR ArduCopter represents the target and executes the trajectory of Bernoulli lemniscate with a focal distance of 20 m. All simulation parameters, such as the rate of the algorithms, image size, range of depth measurements and so on, are set to meet the constraints of on-board processing on the Eagle. The model of the Eagle was designed for Gazebo simulation, while the simulated environment corresponds to the dimensions of the Challenge 1 arena. In order to simulate signals and features of the actual autopilot running on the embedded hardware of the Eagle UAV, the software in the loop (SITL) simulation was set up. Using the SITL simulation, the behaviour of software components can be tested as in the real-world application, making it easy to verify new features and the stability of complex systems.

For the target speed ranging from 2 to 6 m/s, the results of the fully autonomous missions are presented in Figure 10. In both boxplots presented, the central red mark represents the median and the boundaries of the box represent the 25th and 75th percentiles. The whiskers indicate the maximum and minimum data points that are not considered outliers, and outliers are presented with a plus sign. The results are obtained from a total of 75 experiments, meaning 15 trials per each value of the target speed. The success rate of all experiments is 100%, meaning that the point of interception was successfully estimated in less than 3 loops of the target trajectory. For the first four speed values, the states of the Follow-and-Intercept state machine were sequentially executed in all experiments. As for the experiments for the target speed of 6 m/s, the 80% of the trials were also performed sequentially, while in the remaining ones the interruptions of the FOLLOW mode occurred (the follower lost the target) and the SEARCH mode was activated. At higher speeds than the ones presented, the more interruptions occur and the success of finding a suitable interception point cannot be guaranteed.

The maximum achievable speed of the follower while in FOLLOW mode is 4.5 m/s, due to limitations of the control algorithm, which is designed for more general purposes and is not the best choice for visual servoing at high speeds. The boxplot of the Euclidean distance between the estimated interception point and the closest point on the target trajectory is shown in Figure 10a. In cases where the target speed is below the value of the maximum follower speed, the interception point is estimated very accurately and the median of Euclidean distance to the target is less than 0.3 m. If the target is faster than the follower, the depth measurements may fall outside the measurable range, which is the main reason for the increase in the Euclidean distance value for higher speeds, such as 5 and 6 m/s. However, for 75% of the trials at a speed of 5 m/s and more than 50% of the trials at a speed of 6 m/s, the error of the estimated interception point is less than 1.25 m, which means that no additional visual servoing is required. In the remaining trials, visual servoing can eliminate the error in the final approach of a target approaching on the straight part of the figure-eight. As can be seen in Figure 10b, the estimation of the focal distance of Bernoulli lemniscate is generally very accurate, and positive error values indicate a tendency to expand the size of the lemniscate. A different behaviour is obtained in the experiments at a target's speed of 6 m/s (30% faster than the interceptor), where the size of the lemniscate is underestimated due to the scarce observations of the target. In the first iterations of the lemniscate estimation, the value of the bidirectional Hausdorff distance increases linearly with the number of obtained measurements, as shown in Figure 10c from the experiment with the target speed of 5 m/s. As the estimation of focal distance converges, the lemniscate estimation approaches the true trajectory and the value of the Hausdorff distance decreases. The trajectory estimation is deemed to be successful (the green area in the graph) when the condition in step 5 of Algorithm 2 is satisfied. The value of the threshold is a hyperparameter of the trajectory estimation.

**Table 2.** Estimation of the Bernoulli lemniscate

| Run | 1 | 2 | 3 | 4 | 5 | 6 |
|---|---|---|---|---|---|---|
| Focal distance $a$ | 10.578339 | 9.698894 | 9.338982 | 9.452426 | 10.360296 | 10.044878 |
| Euclidean distance | 1.081427 | 0.255335 | 0.868916 | 1.310884 | 0.550260 | 0.184534 |

| Run | 7 | 8 | 9 | 10 | 11 | 12 |
|---|---|---|---|---|---|---|
| Focal distance $a$ | 9.388806 | 9.579864 | 9.832463 | 10.580500 | 9.570580 | 9.038918 |
| Euclidean distance | 0.611240 | 1.268653 | 1.080753 | 0.416507 | 1.845678 | 0.088793 |

### 7.4. Evaluation of the interception in field experiments

Finally, we emulated Challenge 1 in field experiments at our test site, shown in Figures 4 and 5. With simulation experiments validating the visual servoing algorithm and the state machine, the focus of the field experiments was on the ability to intercept the UAV by collecting enough measurements to correctly identify the interception point in real world conditions. For safety reasons[2] no interception was attempted during these experiments and the speed of the target UAV was limited to 1 m/s. With this limit introduced, the intercepting UAV is always capable of following the target (limiting interceptor's speed would require a complete overhaul of the low-level controllers). The target trajectory is generated as a Bernoulli lemniscate with the focal distance of 10 m.

In Table 2, the estimated focal distance of the lemniscate and the error of the estimated interception point in the form of the Euclidean distance are shown for the 12 real-world experiments. Integrating localization data from both UAVs is performed using GPS measurements, and the position of target in the global coordinate system of the interceptor is used as the ground truth. By analysing all the collected data from the experiments, the few false positive detections and occasional drops in the depth measurements are identified. Despite that, the error of the estimated focal distance of the Bernoulli lemniscate is below 1 m (mean absolute error of 0.47 m). As for the interception point, the majority (75 %) of the experiments do not require additional servoing (average distance to the target trajectory of 0.8 m).

The visual representation of the estimated Bernoulli lemniscate (marked pink) compared to the ground truth of the target trajectory (marked red) is shown in the Figure 11. The yellow points are the raw measurements and the green arrow represents the position and orientation of the interceptor UAV at the estimated interception point. In the run 6, which is shown on the left image, a good amount of measurements is obtained in one loop of the trajectory and the estimated Bernoulli lemniscate corresponds well to the ground truth trajectory, which is the desired outcome of the our strategy. The center image shows a case (Run 2) where data on only half of the lemniscate figure is collected, but the estimated interception point is still on the intruder's trajectory and looking in the direction of the incoming intruder. Although more measurements generally mean a better estimate, the uneven distribution of measurements results in shifted estimation. This situation occurred in Run 11, where the largest error of the interception point was obtained. Occasional outliers do not affect the trajectory estimation, as can be seen in the figure.

---

[2] Both the target and the interceptor are Kopterworx Eagle UAVs, which are large rotorcrafts by today's standards (1.13 m motor-to-motor with 0.56 m propellers).

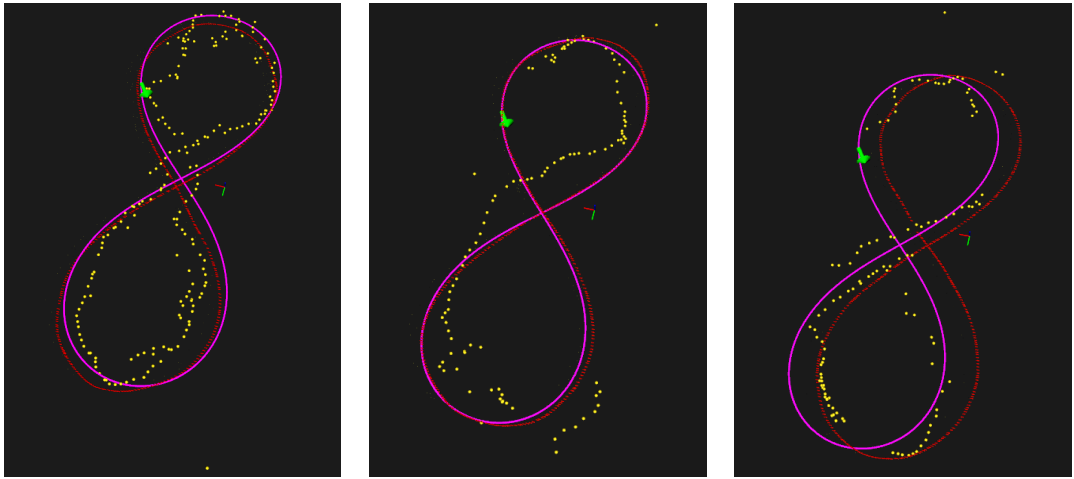

**Figure 11.** The results of the field experiments. The images from left to right correspond to experiments 6, 2 and 11, respectively. The target executes the trajectory (marked in red) of the Bernoulli lemniscate with the focal distance of 10 m. The collected measurements (yellow points) of the target position are used to determine the estimated lemniscate (marked in pink). The successfully estimated interception point (green arrow) lies on the target trajectory pointing in the direction of the incoming target.

Video showcasing the experiments presented in this paper is available at https://www.youtube .com/watch?v=EPoxrC6S8tw.

## 8. Conclusion

In this paper we present a generalized solution to the MBZIRC 2020 Challenge 1. An overview of the hardware and software components, specifically designed for the competition, is given. The motivation for our work, which also applies to the Challenge 1, lies in the safety of UAVs.

The competition performance and experimental results demonstrate our efficient, visual-perception module. The system can extract target UAV trajectory in detail sufficient to enable interception. Based on *a priori* knowledge of the shape of the target trajectory, we managed to track and intercept an intruding drone 30% faster than our sentry vehicle in more than half of the conducted SITL experiments. In the outdoor unstructured environment, we tested system components that analyze the target's behavior and estimate its trajectory — so generating candidate interception points — and achieved successful intercepts in 9 of 12 experiments.

In the future work, the development of the proposed system will be continued. At this point, the improvement in the control algorithms, the increase of the detection rate and the introduction of a family of target trajectory shapes are considered.

## Acknowledgments

Research work presented in this paper has been supported by Khalifa University of Science and Technology, Abu Dhabi, UAE, and by European Commission Horizon 2020 Programme through project under G. A. number 810321, named Twinning coordination action for spreading excellence in Aerial Robotics - AeRoTwin. Work of Frano Petric was supported by the European Regional Development Fund under the grant KK.01.1.1.01.0009 (DATACROSS).

## ORCID

Antonella Barišić  https://orcid.org/0000-0002-4532-6915
Frano Petric  https://orcid.org/0000-0002-9806-8569
Stjepan Bogdan  https://orcid.org/0000-0003-2636-3216

## References

Abughalwa, M., Samara, L., Hasna, M. O., and Hamila, R. (2020). Full-duplex jamming and interception analysis of UAV-based intrusion links. *IEEE Communications Letters*, 24(5):1105–1109.

AeroGuard (2020). Autonomous, rapid-response drone interdiction to stop the 3d threat. Available: https://www.sci.com/aeroguard/, Accessed: 3 March 2021.

Barisic, A., Car, M., and Bogdan, S. (2019). Vision-based system for a real-time detection and following of UAV. In *2019 Workshop on Research, Education and Development of Unmanned Aerial Systems (RED UAS)*, pages 156–159. IEEE.

Beard, R., McLain, T., Goodrich, M., and Anderson, E. (2002). Coordinated target assignment and intercept for unmanned air vehicles. *IEEE Transactions on Robotics and Automation*, 18(6):911–922.

Best, K. L., Schmid, J., Tierney, S., Awan, J., Beyene, N. M., Holliday, M. A., Khan, R., and Lee, K. (2020). *How to Analyze the Cyber Threat from Drones*. RAND Corporation.

Beul, M., Nieuwenhuisen, M., Quenzel, J., Rosu, R. A., Horn, J., Pavlichenko, D., Houben, S., and Behnke, S. (2019). Team nimbro at mbzirc 2017: Fast landing on a moving target and treasure hunting with a team of micro aerial vehicles. *Journal of Field Robotics*, 36(1):204–229.

Bond, E., Crowther, B., and Parslew, B. (2019). The rise of high-performance multi-rotor unmanned aerial vehicles - how worried should we be? In *2019 Workshop on Research, Education and Development of Unmanned Aerial Systems (RED UAS)*, pages 177–184. IEEE.

Brust, M. R., Danoy, G., Stolfi, D. H., and Bouvry, P. (2021). Swarm-based counter UAV defense system. *Discover Internet of Things*, 1(2).

Chaumette, F. and Hutchinson, S. (2006). Visual servo control. i. basic approaches. *IEEE Robotics Automation Magazine*, 13(4):82–90.

Császár, Á. (1978). *General Topology*. Disquisitiones mathematicae Hungaricae. Akadémiai Kiadó.

Duan, K., Bai, S., Xie, L., Qi, H., Huang, Q., and Tian, Q. (2019). CenterNet: Keypoint triplets for object detection. In *2019 IEEE/CVF International Conference on Computer Vision (ICCV)*, pages 6568–6577. IEEE.

Dumitrescu, C., Minea, M., Costea, I. M., Chiva, I. C., and Semenescu, A. (2020). Development of an acoustic system for uav detection. *Sensors*, 20(4870):2–27.

Ezuma, M., Erden, F., Anjinappa, C. K., Ozdemir, O., and Guvenc, I. (2019). Detection and classification of uavs using rf fingerprints in the presence of interference. *IEEE Open Journal of the Communications Society*, 1:60 – 76.

Fox, S. J. (2019). Policing: Monitoring, investigating and prosecuting 'drones'. *European Journal of Comparative Law and Governance*, 6(1):78–126.

Guvenç, I., Koohifar, F., Singh, S., Sichitiu, M. L., and Matolak, D. (2018). Detection, tracking, and interdiction for amateur drones. *IEEE Communications Magazine*, pages 75–81.

Hehn, M. and D'Andrea, R. (2012). Real-time trajectory generation for interception maneuvers with quadcopters. In *2012 IEEE/RSJ International Conference on Intelligent Robots and Systems*, pages 4979–4984. IEEE.

Li, Z., Meng, C., Zhou, F., Ding, X., Wang, X., Zhang, H., Guo, P., and Meng, X. (2018). Fast vision-based autonomous detection of moving cooperative target for unmanned aerial vehicle landing. *Journal of Field Robotics*, 36(1):34–48.

Liu, W., Anguelov, D., Erhan, D., Szegedy, C., Reed, S., Fu, C.-Y., and Berg, A. C. (2016). SSD: Single shot MultiBox detector. In *Computer Vision – ECCV 2016*, pages 21–37. Springer International Publishing.

MBZIRC (2020). The challenge 2020. Available: https://www.mbzirc.com/challenge/2020, Accessed: 30 September 2020.

Moreira, M., Papp, E., and Ventura, R. (2019). Interception of non-cooperative UAVs. In *2019 IEEE International Symposium on Safety, Security, and Rescue Robotics (SSRR)*, pages 120–125. IEEE.

Pearson, K. (1901). Liii. on lines and planes of closest fit to systems of points in space. *The London, Edinburgh, and Dublin Philosophical Magazine and Journal of Science*, 2(11):559–572.

Pham, H. and Pham, Q. (2018). A new approach to time-optimal path parameterization based on reachability analysis. *IEEE Transactions on Robotics*, 34(3):645–659.

Puljiz, D., Varga, M., and Bogdan, S. (2012). Stochastic search strategies in 2d using agents with limited perception. *IFAC Proceedings Volumes*, 45(22):650–654.

Redmon, J. and Farhadi, A. (2018). Yolov3: An incremental improvement. *CoRR*, abs/1804.02767.

Sinopoli, B., Micheli, M., Donato, G., and Koo, T. J. (2001). Vision based navigation for an unmanned aerial vehicle. In *Proceedings 2001 ICRA. IEEE International Conference on Robotics and Automation (Cat. No.01CH37164)*, volume 2, pages 1757–1764 vol.2.

Tzoumanikas, D., Li, W., Grimm, M., Zhang, K., Kovac, M., and Leutenegger, S. (2018). Fully autonomous micro air vehicle flight and landing on a moving target using visual-inertial estimation and model-predictive control. *Journal of Field Robotics*, 36(1):49–77.

Vrba, M., Hert, D., and Saska, M. (2019). Onboard marker-less detection and localization of non-cooperating drones for their safe interception by an autonomous aerial system. *IEEE Robotics and Automation Letters*, 4(4):3402–3409.

Vrba, M. and Saska, M. (2020). Marker-less micro aerial vehicle detection and localization using convolutional neural networks. *IEEE Robotics and Automation Letters*, 5(2):2459–2466.

Yaacoub, J.-P., Noura, H., Salman, O., and Chehab, A. (2020). Security analysis of drones systems: Attacks, limitations, and recommendations. *Internet of Things*, 11:100218.

Yang, K. and Quan, Q. (2020). An autonomous intercept drone with image-based visual servo. In *2020 IEEE International Conference on Robotics and Automation (ICRA)*, pages 2230–2236. IEEE.

Zhu, C., He, Y., and Savvides, M. (2019). Feature selective anchor-free module for single-shot object detection. In *2019 IEEE/CVF Conference on Computer Vision and Pattern Recognition (CVPR)*, pages 840–849. IEEE.

Çetin, E., Barrado, C., and Pastor, E. (2020). Counter a drone in a complex neighborhood area by deep reinforcement learning. *Sensors*, 20(2320):1–25.